\documentclass[11pt]{article}

\usepackage[]{acl}
\usepackage{times}
\usepackage{latexsym}
\usepackage[T1]{fontenc}
\usepackage[utf8]{inputenc}
\usepackage{microtype}
\usepackage{inconsolata}
\usepackage{graphicx}
\usepackage{amsmath}
\usepackage{booktabs}
\usepackage[ruled,vlined,linesnumbered]{algorithm2e}
\usepackage[font=footnotesize]{caption}
\usepackage[table]{xcolor}
\usepackage{colortbl}
\usepackage{setspace}
\title{CREST: Curvature-Regulated Event-Centric Sampling for
Efficient Long-Video Understanding}

\author{
  Mehrajul Abadin Miraj$^1$ \quad
  Abdul Mohaimen Al Radi$^2$ \quad
  Shariful Islam Rayhan$^1$ \quad
  Md. Tanvir Alam$^1$ \\
  Ismat Rahman$^1$ \quad
  Yu Tian$^2$ \quad
  Md Mosaddek Khan$^1$ \\[0.4em]
  $^1$Department of Computer Science and Engineering, University of Dhaka \\
  $^2$Department of Computer Science and Engineering, University of Central Florida \\[0.3em]
  \small\texttt{mehrajulabadin-2020115630@cs.du.ac.bd} \quad
  \small\texttt{ab575577@ucf.edu}
}
\author{
  Mehrajul Abadin Miraj$^{1\dagger}$ \\
  \And
  Abdul Mohaimen Al Radi$^{2\ddagger}$ \\
  \And
  Shariful Islam Rayhan$^1$ \\
  \AND
  Md. Tanvir Alam$^1$ \\
  \And
  Ismat Rahman$^1$ \\
  \And
  Yu Tian$^2$ \\
  \AND
  Md Mosaddek Khan$^{1\S}$ \\
  $^1$Dept. of CSE, University of Dhaka \quad
  $^2$Dept. of CSE, University of Central Florida \\
  $^\dagger$\texttt{mehrajulabadin-2020115630@cs.du.ac.bd} \quad
  $^\ddagger$\texttt{ab575577@ucf.edu} \quad
  $^\S$\texttt{mosaddek@du.ac.bd}
}
\author{
  Mehrajul Abadin Miraj$^{1,a}$ \\
  \And
  Abdul Mohaimen Al Radi$^{2,b}$ \\
  \And
  Shariful Islam Rayhan$^{1,c}$ \\
  \AND
  Md. Tanvir Alam$^{1,d}$ \\
  \And
  Ismat Rahman$^{1,e}$ \\
  \And
  Yu Tian$^{2,f}$ \\
  \AND
  Md Mosaddek Khan$^{1,g}$ \\
  $^1$Dept. of CSE, University of Dhaka \quad
  $^2$Dept. of CSE, University of Central Florida \\
  $^a$\texttt{mehrajulabadin-2020115630@cs.du.ac.bd} \quad
  $^b$\texttt{ab575577@ucf.edu} \\
  $^c$\texttt{mdsharifulislam-2020815651@cs.du.ac.bd} \quad
  $^d$\texttt{tanvir15@du.ac.bd} \\
  $^e$\texttt{ismat@cse.du.ac.bd} \quad
  $^f$\texttt{yu.tian2@ucf.edu} \quad
  $^g$\texttt{mosaddek@du.ac.bd}
}

\begin{document}
\maketitle
\begin{abstract}
Selecting informative frames from long videos is a combinatorial problem that existing
methods address either through efficient heuristics without explicit modeling of
query-conditioned temporal structure, or through multi-stage retrieval pipelines with
substantial preprocessing cost. We propose \textbf{CREST}, a training-free frame
selection method grounded in the temporal geometry of query--frame relevance. CREST is
based on the observation that relevance over time exhibits structured local variation:
sharp curvature around salient events and flatter regions in redundant segments. By
using local curvature to guide selection, CREST allocates a fixed frame budget more
effectively across brief decisive events and slowly evolving evidence. Under a fixed
backbone and frame budget, CREST achieves higher accuracy than AKS, a lightweight
relevance--coverage baseline, on LongVideoBench and VideoMME, while retaining
93--95\% of the accuracy of MIRA, a stronger multi-stage retrieval pipeline, at only
3--4\% of its preprocessing cost.\footnote{Code and implementation details are included
in the supplementary material and will be released publicly upon acceptance.} On
TempRel, our diagnostic benchmark for temporal frame selection, CREST achieves a
6.88\% relative improvement over AKS. Pairwise LLM-as-a-judge evaluation further shows
that CREST-selected frames yield more coherent frame-conditioned descriptions, with win
rates of 60.58\% and 54.50\% on the two benchmarks. These results show that local
temporal geometry provides a simple and efficient basis for long-video frame selection.
\end{abstract}

\begin{figure}[t]
    \centering
    \includegraphics[width=1\linewidth]{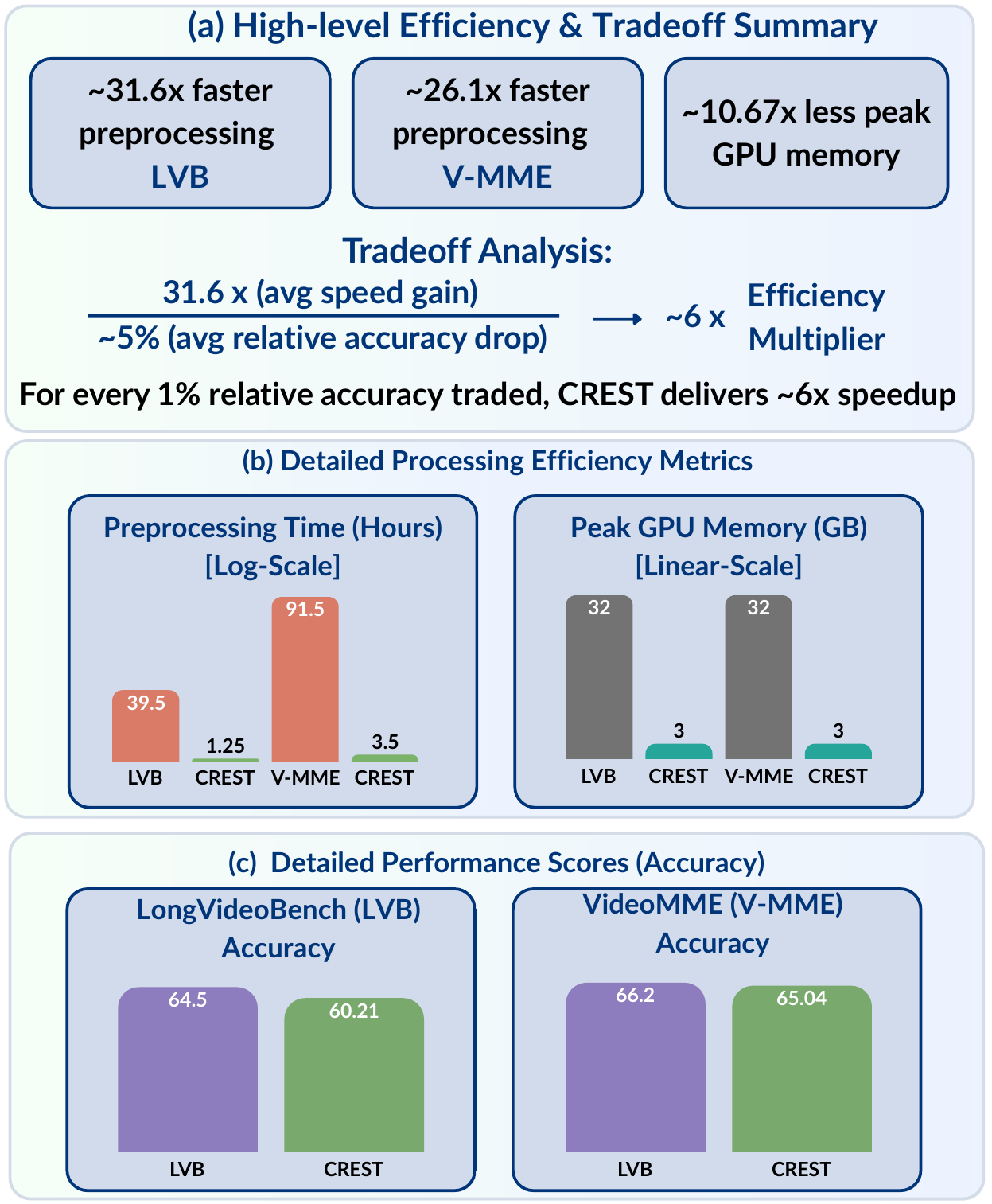}
    \captionsetup{font=footnotesize}
    \caption{Efficiency--performance trade-off of CREST compared to MIRA. CREST reduces
    preprocessing time by up to ${\sim}31.6\times$ and memory usage by
    ${\sim}10.7\times$, while retaining 93--95\% of MIRA's accuracy. Effective frame
    selection can achieve near-equivalent performance at a fraction of the computational
    cost.}
    \label{fig:efficiency_accuracy_tradeoff}
\end{figure}

\section{Introduction}
Long-video understanding requires selecting, from potentially thousands of frames, the
small subset of visual evidence needed to answer a natural-language query under a fixed
frame budget. Selecting $M$ frames from $T$ candidates is a combinatorial problem, while
modern multimodal large language models (MLLMs) can process only a limited number of
visual tokens, making exhaustive frame encoding computationally impractical for long
videos~\citep{bai2023qwenvl, chen2023internvl, liu2023llava, zhu2023minigpt4,
tang2025aks}. The frame-selection policy therefore determines both computational
efficiency and whether the downstream model receives the temporally localized evidence
needed for accurate and grounded reasoning.

\begin{figure*}[t]
    \centering
    \includegraphics[width=1\linewidth]{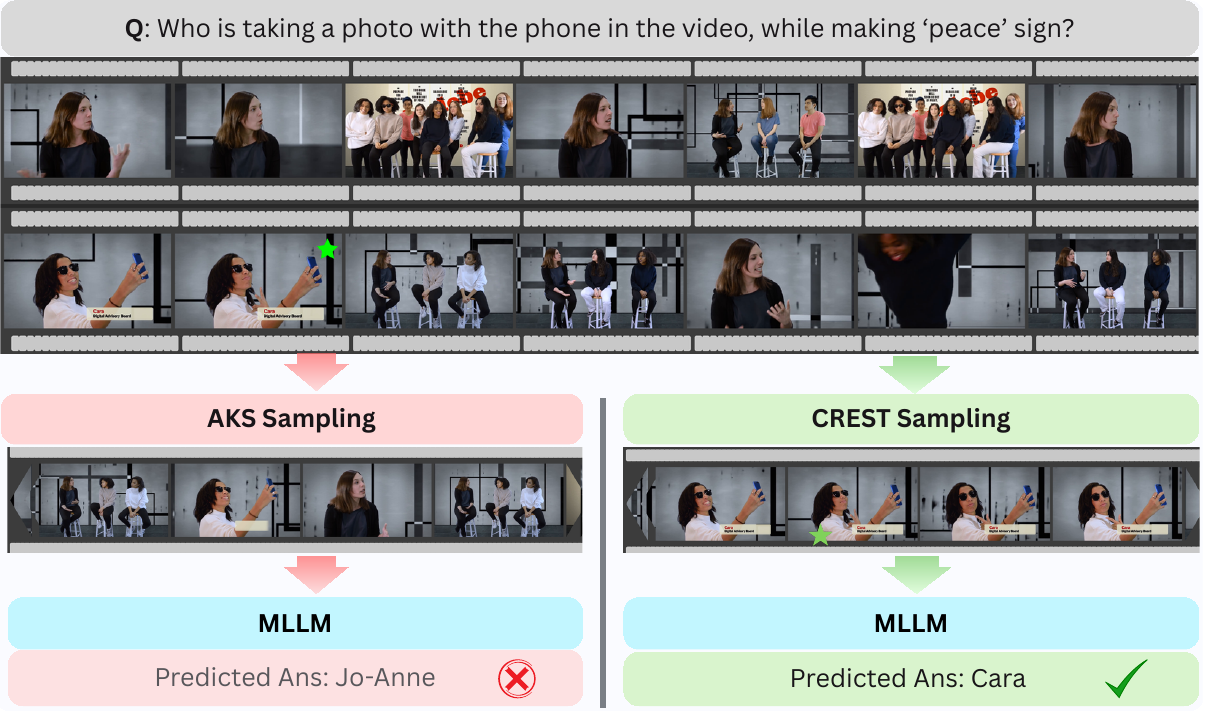}
    \caption{Comparison of keyframe selection strategies. Given the same query, AKS
    distributes selections based on a relevance--coverage trade-off and fails to capture
    the critical moment, leading to an incorrect prediction. CREST concentrates
    selections around regions of high curvature in the relevance signal, recovering the
    key event and enabling the correct answer.}
    \label{fig:selection_comparison}
\end{figure*}

% Single-column figure: [t] is valid here

This constrained setting has led to a range of frame-selection strategies. Simple
heuristics such as uniform sampling~\citep{zhang2024llavavideo} and top-$K$
relevance selection~\citep{tang2025aks} are efficient, but they treat frames largely as
isolated candidates and can miss brief, query-relevant events. More structured methods
introduce explicit coverage or retrieval mechanisms: AKS~\citep{tang2025aks}
recursively allocates keyframes through a relevance--coverage trade-off, while
MIRA~\citep{hao2026mira} uses multi-view relevance scoring and adaptive routing to
achieve stronger performance at substantially higher preprocessing cost. These methods
improve over naive sampling, but they do not explicitly model the local temporal
geometry of query-conditioned relevance.

This gap is consequential because the query-conditioned relevance signal is highly
non-uniform over time. Informative content may appear as sharp peaks around brief
decisive events or as broad plateaus across slowly evolving, redundant regions. Treating
these regimes identically can either miss critical moments or waste the frame budget on
visually similar evidence. As illustrated in Figure~\ref{fig:selection_comparison}, a
relevance--coverage allocation can spread selections away from the decisive event,
whereas modeling the local shape of the relevance signal can recover the event region
required by the query.

We propose \textbf{CREST} (Curvature-Regulated Event-Centric Sampling), a lightweight,
training-free frame-selection method for query-conditioned long-video understanding.
CREST treats query--frame relevance scores as a temporal signal and uses local curvature
to regulate non-maximum suppression. Around high-curvature peaks, the suppression radius
contracts to preserve dense evidence near brief decisive events; in flatter regions, it
remains broader to reduce redundant selections from slowly varying content. A temporal
decay mechanism further relaxes earlier suppression decisions as selection proceeds,
allowing informative neighboring regions to re-enter consideration under the fixed frame
budget. As shown in Figure~\ref{fig:efficiency_accuracy_tradeoff}, this simple
geometric rule yields a favorable efficiency--accuracy trade-off, retaining 93--95\% of
MIRA's accuracy while requiring only 3--4\% of its preprocessing cost. Our contributions
are threefold:

\begin{enumerate}

\item \textbf{Temporal-geometry-aware frame selection.} We formulate query-conditioned
frame selection through the local geometry of the relevance signal, introducing a
training-free selection rule that goes beyond score magnitude and global coverage.

\item \textbf{Efficient long-video reasoning.} Under a shared backbone and fixed frame
budget, CREST achieves higher accuracy than AKS on LongVideoBench and VideoMME while
offering a substantially lower-cost alternative to MIRA
\citep{wu2024longvideobench, fu2024videomme}.

\item \textbf{Diagnostic and evidence-oriented evaluation.} We introduce TempRel to test
frame selection under extended and context-dependent temporal relevance regimes, and
complement multiple-choice accuracy with a pairwise description-based evaluation of
whether selected frames support richer frame-conditioned descriptions.

\end{enumerate}

% \section{Related Works}
% See Appendix x for related works.

\section{Methodology}
\label{sec:methods}
We address query-conditioned frame selection for long-video understanding: given a
video and a natural-language query, the goal is to select $M$ frames that provide useful
visual evidence under a fixed frame budget. Section~\ref{subsec:Preliminaries}
formalizes this as a constrained combinatorial selection problem and defines the
query--frame relevance proxy. Section~\ref{subsec:principles} motivates a
temporal-geometric view of this proxy, and Section~\ref{subsec:crest} presents CREST as
a greedy curvature-regulated procedure with initialization, curvature-adaptive
suppression, and temporal decay.

\subsection{Preliminaries}
\label{subsec:Preliminaries}

We consider video question answering, where a model receives a video $\mathbf{V}$ and a
textual query $\mathbf{Q}$ and must produce a textual response. The video is represented
as an ordered sequence of frames $\mathbf{V}=\{F_1,F_2,\dots,F_T\}$, where $T$ is the
total number of frames. Since MLLMs encode frames as visual tokens and operate under a
fixed context budget, only $M \ll T$ frames can be passed to the downstream model. The
task is therefore to select an index set
$\mathcal{I}\subset\{1,\dots,T\}$ with $|\mathcal{I}|=M$ that maximizes
query-dependent utility:
\[
    \mathcal{I}^* = \arg\max_{|\mathcal{I}| = M}
    \mathcal{U}\!\left(\{F_t \mid t \in \mathcal{I}\},\, \mathbf{Q}\right).
\]
Directly optimizing this objective is infeasible because the number of candidate subsets
is $\binom{T}{M}$ and no frame-level supervision is available for the target utility.
We therefore approximate the objective using query--frame relevance as a proxy signal.

\begin{algorithm}[t]
\DontPrintSemicolon
\SetAlgoLined
\caption{CREST: Curvature-Regulated Event-Centric Sampling}
\label{alg:crest}
\KwIn{Relevance scores $\{s_t\}_{t=1}^{T}$, frame budget $M$, decay parameter $\rho$}
\KwOut{Selected frame indices $\mathcal{I}$}
Normalize $s_t \leftarrow s_t / \max_t s_t$\;
Compute $\kappa_t = |s_{t+1} - 2s_t + s_{t-1}|$ for all $t$\;
$R_{\text{base}} \leftarrow T/M$\;
$\lambda \leftarrow \ln 2\,/\,(\rho M)$;\quad
$\mathcal{I} \leftarrow \emptyset$;\quad $n \leftarrow 0$\;
\While{$|\mathcal{I}| < M$}{
    $i^* \leftarrow \arg\max_t\, s_t$\;
    $\mathcal{I} \leftarrow \mathcal{I} \cup \{i^*\}$;\quad $n \leftarrow n + 1$\;
    $R_{i^*} \leftarrow R_{\text{base}}\,/\,(1 + \kappa_{i^*})$\;
    $s_t \leftarrow 0$ \textbf{ for all } $t \in \mathcal{N}(i^*,\, R_{i^*})$\;
    \For{$j \in \mathcal{I} \setminus \{i^*\}$}{
        $\Delta s \leftarrow n - n_j$
        \tcp*{steps since $j$ was selected}
        $R_j^{(n)} \leftarrow R_j \exp(-\lambda\,\Delta s)$\;
        restore $s_t$ for $t$ newly outside $\mathcal{N}(j,\, R_j^{(n)})$\;
    }
}
\Return{$\mathcal{I}$}
\end{algorithm}
\setlength{\textfloatsep}{6pt plus 1pt minus 2pt}

\begin{figure*}[t]
    \centering
    \includegraphics[width=1\linewidth]{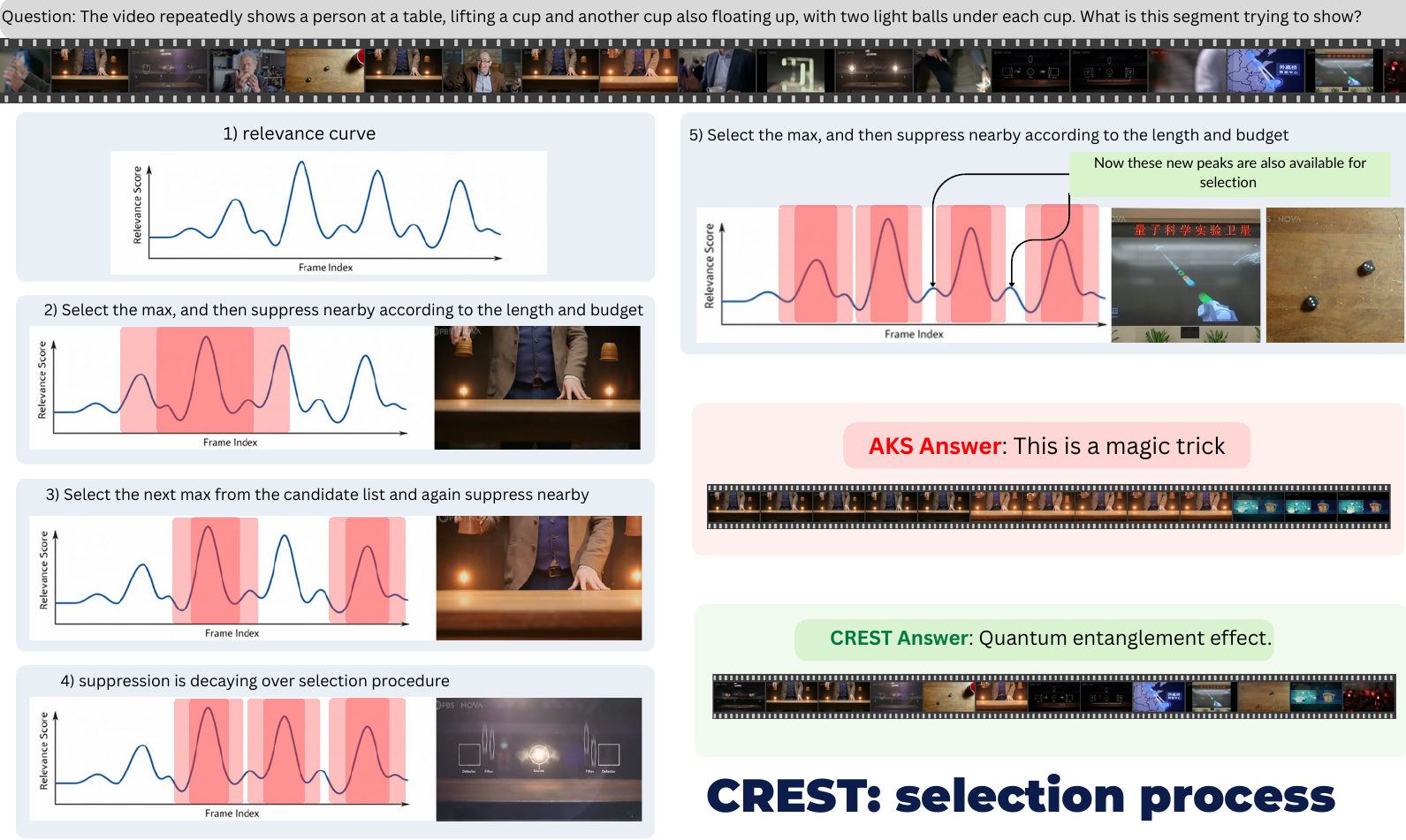}
    \caption{Illustration of the CREST selection process. At each iteration, the
    highest-scoring frame is selected and a curvature-modulated suppression is applied
    to its neighborhood: the radius contracts around sharp peaks and expands in flatter
    regions. As selection proceeds, suppression from earlier iterations decays, allowing
    previously suppressed regions to re-emerge as candidates.}
    \label{fig:selection_algorithm}
\end{figure*}

\subsection{From Score Magnitude to Temporal Geometry}
\label{subsec:principles}
A natural proxy for query-dependent utility is query--frame relevance. Let
$s_t=s(\mathbf{Q},F_t)$ denote the relevance score between the query $\mathbf{Q}$ and
frame $F_t$, computed using a pretrained vision--language model such as CLIP without
additional training. As illustrated in Figure~\ref{fig:selection_algorithm} (panel~1), Figure~\ref{fig:verifiability_event}
(panel~2), the sequence $\{s_t\}_{t=1}^{T}$ forms a highly non-uniform temporal signal:
sharp, narrow peaks at critical events and extended plateaus where relevance evolves
gradually. Selecting the top-$K$ highest-scoring frames is simple but produces redundant
selections, since temporally adjacent frames share similar content. Prior
methods~\citep{tang2025aks} address this by adding diversity or coverage constraints,
improving temporal spread while still not explicitly modeling the local shape of the
relevance signal.

The underlying issue is that score magnitude alone cannot distinguish these two
regimes. A high score at the crest of a sharp peak signals a brief, semantically dense
event; the same score on a broad plateau signals gradual, redundant content. Effective
frame selection requires modeling the \textit{temporal geometry} of the
query-conditioned relevance signal, not just its pointwise values.

\subsection{CREST: Curvature-Regulated Event-Centric Sampling}
\label{subsec:crest}

CREST instantiates the temporal-geometric view as a greedy, iterative selection
procedure. The algorithm maintains a working copy of the relevance scores and proceeds
in rounds: at each step, it selects the highest-scoring remaining frame and suppresses a
temporal neighborhood around it so that subsequent iterations are encouraged to select
complementary evidence. Unlike fixed-window non-maximum suppression, CREST adapts the
suppression radius to local curvature: high-curvature peaks receive narrower
suppression to preserve dense evidence around brief events, while flatter regions
receive broader suppression to reduce redundant selections. The procedure consists of
three steps.

\medskip
\medskip
\noindent\textbf{Step 1: Initialization.}
Before selection begins, scores are normalized to $[0,1]$ and local curvature is
computed over the relevance sequence (Algorithm~\ref{alg:crest}, lines~1--3). We
estimate curvature using a discrete second-order finite difference:
\[
    \kappa_t = \left| s_{t+1} - 2s_t + s_{t-1} \right|,
\]
with boundary frames handled by one-sided differences. This quantity is large when the
relevance signal changes sharply around frame $t$ and small when the signal is locally
flat, making it a lightweight proxy for local temporal variation without requiring any
learned components. We set the baseline suppression radius as
\[
    R_{\text{base}} = \frac{T}{M},
\]
which corresponds to the average spacing obtained by uniformly distributing $M$
selections across $T$ frames. This radius provides the scale that curvature modulates.

\medskip
\noindent\textbf{Step 2: Greedy selection with curvature-adaptive suppression.}
The main loop (Algorithm~\ref{alg:crest}, lines~5--9) iterates until the frame budget
is exhausted. At each iteration, CREST selects the highest-scoring remaining frame and
assigns it a curvature-regulated suppression radius:

\[
    R_{i^*} = \frac{R_{\text{base}}}{1 + \kappa_{i^*}}.
\]
High curvature at $i^*$, which signals a sharp, brief event, contracts the radius,
so that only the immediate vicinity is suppressed and nearby frames remain available
as candidates. Low curvature, which signals a broad, slowly varying region, keeps the
radius closer to the baseline spacing, suppressing a wider neighborhood and preventing
the budget from being consumed by redundant frames. All scores within the neighborhood
$\mathcal{N}(i^*, R_{i^*}) = \{t \mid |t - i^*| \leq R_{i^*}\}$ are then zeroed
(line~9):
\[
    s_t \leftarrow 0, \quad \forall\, t \in \mathcal{N}(i^*,\, R_{i^*}).
\]
This operation implements non-maximum suppression with a curvature-varying window. As
shown in Figure~\ref{fig:selection_algorithm} (panel~2), the selected peak and its
highlighted suppression band are removed from the candidate set, exposing the next
informative region. The following iteration (panel~3) then selects the next dominant
peak from the remaining signal, and the process repeats.

\medskip

\begin{table*}[t]
\centering
\caption{Video question answering accuracy (\%) on LongVideoBench (LVB) and
VideoMME (V-MME).}
\label{tab:overall_comparison}
\resizebox{\linewidth}{!}{%
\begin{tabular}{llcccc}
\toprule
Method & Type & Model & Frames & LVB val & V-MME \\
\midrule
\multicolumn{6}{l}{\footnotesize Proprietary model results as reported in AKS~\cite{tang2025aks}.} \\
\rowcolor{gray!15}
GPT-4V \cite{openai2023gpt4v} & Proprietary & -- & 256 & 61.3 & 59.9 \\
\rowcolor{gray!15}
GPT-4o \cite{openai2024gpt4o} & Proprietary & -- & 256 & 66.7 & 71.9 \\
\rowcolor{gray!15}
Gemini-1.5-Flash \cite{gemini15} & Proprietary & -- & 256 & 61.6 & 70.3 \\
\rowcolor{gray!15}
Gemini-1.5-Pro \cite{gemini15} & Proprietary & -- & 256 & 64.0 & 75.0 \\
\midrule
Video-LLaVA \citep{lin2023videollava} & Foundational & Video-LLaVA-7B & 8 & 39.1 & 39.9 \\
MiniCPM-V 2.6 \cite{minicpmv2024} & Foundational & MiniCPM-V 2.6-8B & 64 & 54.9 & 60.9 \\
PLLaVA \citep{xu2024pllava} & Foundational & PLLaVA-34B & 32 & 53.2 & -- \\
VILA \citep{liu2024vila} & Foundational & VILA-40B & -- & -- & 60.1 \\
LLaVA-Video \citep{zhang2024llavavideo} & Foundational
    & LLaVA-Video-7B & 64 & 58.2 & 63.33 \\
\midrule
LongVILA \cite{chen2024longvila} & Training-based & LongVILA-7B & 256 & 57.1 & 60.1 \\
LongVU \cite{shen2024longvu} & Training-based & LongVU-7B & 1FPS & -- & 60.6 \\
Apollo \cite{zohar2025apollo} & Training-based & Apollo-7B & 2FPS & 58.5 & 61.3 \\
BIMBA \cite{bimba2025} & Training-based & BIMBA-7B & 128 & 59.5 & 64.7 \\
TPO \cite{li2025tpo} & Training-based & LLaVA-Video-7B & 64 & 60.1 & 65.6 \\
\midrule
DToMA \cite{yuan2025dtoma} & Training-free & LLaVA-Video-7B & 64 & 59.6 & 65.0 \\
T* \citep{tstar2025} & Training-free & LLaVA-OneVision-72B & 32 & 65.4 & 68.3 \\
AKS$^\dagger$ \cite{tang2025aks} & Training-free & LLaVA-Video-7B & 32 & 59.76 & 64.48 \\
MIRA \citep{hao2026mira} & Training-free & LLaVA-Video-7B & 64 & 64.5 & 66.2 \\
\textbf{CREST (ours)$^\dagger$} & Training-free & LLaVA-Video-7B & 32
    & \textbf{60.21} & \textbf{65.04} \\
\bottomrule
\end{tabular}%
}
\vspace{2pt}
{\raggedright \footnotesize $^\dagger$\,denotes our reproduced experimental results.\par}
\end{table*}

\noindent\textbf{Step 3: Temporal decay for full-video coverage.}
A fixed suppression radius would permanently exclude frames near early selections. To prevent this, the suppression radius of every previously
selected frame decays exponentially with each subsequent selection step
(Algorithm~\ref{alg:crest}, lines~10--12):
\[
    R_i^{(n)} = R_i \exp(-\lambda\, \Delta s), \qquad \lambda = \frac{\ln 2}{\rho M},
\]
where $\Delta s$ is the number of selection steps elapsed since frame $i$ was chosen
and $\rho \in (0,1)$ controls the decay rate. As each radius shrinks, scores in the
formerly suppressed neighborhood are progressively restored, returning those frames to
the candidate pool. Panel~4 of Figure~\ref{fig:selection_algorithm} illustrates this
effect: the suppression bands narrow across iterations. By panel~5, the regions around
early selections have recovered sufficiently that their neighboring peaks re-enter
contention, enabling the algorithm to capture important context near high scored frames. The
full procedure is given in Algorithm~\ref{alg:crest}.

\section{Empirical Evaluation}
For performance analysis, we evaluate CREST on two standard long-video understanding benchmarks: LongVideoBench
(LVB)~\citep{wu2024longvideobench} and VideoMME~\citep{fu2024videomme}, both of which
require reasoning over extended temporal contexts. All experiments use LLaVA-Video-7B as
the backbone with a frame budget of $M = 32$ unless stated otherwise. Preprocessing,
including frame extraction, relevance scoring, and frame selection, is performed on a
single NVIDIA RTX 5090 GPU. We adopt the \texttt{lmms\_eval}
framework~\citep{zhang2025lmmseval} for standardized and reproducible evaluation. We evaluate CREST along three axes: (i) accuracy relative to prior lightweight methods under a shared backbone and fixed frame budget (Section~\ref{subsec:main}), (ii) robustness across different frame budgets and relevance scorers through ablation studies (Section~\ref{subsec:ablation}), and (iii) grounding quality via description generation and LLM-as-a-judge evaluation (Section~\ref{subsec:grounding}). Section~\ref{subsec:temprel} reports results on TempRel, a diagnostic benchmark designed to isolate performance under controlled temporal relevance regimes. 
% Table 1: placed BEFORE \subsection so it floats to the top of the
% results page without competing with body text for space.

\subsection{Long-Video Understanding Performance}
\label{subsec:main}
CREST consistently outperforms prior lightweight, training-free methods under the same
backbone and frame budget. Table~\ref{tab:overall_comparison} reports accuracy across
methods. Compared to AKS, CREST achieves +0.45\% on LongVideoBench and +0.56\% on
VideoMME, demonstrating that explicitly modeling the temporal geometry of the relevance
signal leads to more informative frame selection. Results are deterministic across
multiple runs.

Notably, CREST with a 7B backbone and 32 frames achieves competitive accuracy against
significantly larger or closed-source models, confirming that improved frame selection,
not increased model scale, is the primary driver of performance. Table~\ref{tab:efficiency} compares the preprocessing cost against MIRA. CREST achieves a
26--31$\times$ speedup and more than $10\times$ lower peak memory usage, a direct
consequence of avoiding multi-stage retrieval and iterative scoring pipelines.

% Table 2: 5 cols, fits in one column with \small
\begin{table}[t]
\centering
\small
\setlength{\tabcolsep}{4pt}
\caption{Preprocessing efficiency of CREST versus MIRA. Both methods are
evaluated under the same frame budget to isolate the cost of the frame
selection stage. CREST's preprocessing cost is independent of the frame budget $M$: the dominant 
cost is computing relevance scores over all $T$ frames, which does not vary with 
the number of frames selected.}
\label{tab:efficiency}
\begin{tabular}{lcccc}
\toprule
 & \multicolumn{2}{c}{Time (hours)} & \multicolumn{2}{c}{Peak VRAM (GB)} \\
\cmidrule(lr){2-3} \cmidrule(lr){4-5}
Method & LVB & V-MME & LVB & V-MME \\
\midrule
MIRA         & 39.5               & 91.5               & 32               & 32               \\
CREST (ours) & 1.25               & 3.5                & 3                & 3                \\
\midrule
Speedup      & $\sim$31.6$\times$ & $\sim$26.1$\times$ & $\sim$10.7$\times$ & $\sim$10.7$\times$ \\
\bottomrule
\end{tabular}
\end{table}

\subsection{Ablation and Robustness}
\label{subsec:ablation}

% Tables 3, 4, 5: 7 columns each — must be table* (full text width).
% With table (single col), \resizebox squashes them into ~84 mm;
% with table*, \linewidth = full ~174 mm and they render properly.
\begin{table}[t]
\centering
\small
\setlength{\tabcolsep}{3pt}
\caption{Accuracy (\%) under different frame budgets.}
\label{tab:budget}
\begin{tabular}{lcccccc}
\toprule
& \multicolumn{3}{c}{LongVideoBench} & \multicolumn{3}{c}{VideoMME} \\
\cmidrule(lr){2-4} \cmidrule(lr){5-7}
Method & $k{=}16$ & $k{=}32$ & $\Delta$ & $k{=}16$ & $k{=}32$ & $\Delta$ \\
\midrule
AKS          & 58.41          & 59.76          & +1.35          & 62.88          & 64.48          & +1.60 \\
CREST (ours) & \textbf{59.32} & \textbf{60.21} & \textbf{+0.89} & \textbf{63.91} & \textbf{65.04} & \textbf{+1.13} \\
\bottomrule
\end{tabular}
\end{table}
\begin{table}[t]
\centering
\small
\setlength{\tabcolsep}{3pt}
\caption{Accuracy (\%) across vision--language relevance scorers ($k{=}32$).}
\label{tab:vl_scorer}
\begin{tabular}{lcccccc}
\toprule
& \multicolumn{3}{c}{LongVideoBench} & \multicolumn{3}{c}{VideoMME} \\
\cmidrule(lr){2-4} \cmidrule(lr){5-7}
Method & BLIP & CLIP & Sevila & BLIP & CLIP & Sevila \\
\midrule
AKS          & 59.76          & 58.41          & 58.86          & 64.48          & 64.33          & 63.56 \\
CREST (ours) & \textbf{60.21} & \textbf{59.88} & \textbf{59.16} & \textbf{65.04} & \textbf{65.04} & \textbf{64.19} \\
\bottomrule
\end{tabular}
\end{table}
\begin{table}[t]
\centering
\small
\setlength{\tabcolsep}{3pt}
\caption{Ablation study on LongVideoBench and VideoMME ($k{=}32$).}
\label{tab:ablation}
\begin{tabular}{lcccccc}
\toprule
& \multicolumn{3}{c}{LongVideoBench} & \multicolumn{3}{c}{VideoMME} \\
\cmidrule(lr){2-4} \cmidrule(lr){5-7}
Variant & BLIP & CLIP & Sevila & BLIP & CLIP & Sevila \\
\midrule
w/o curvature & 59.31          & 58.12          & 58.79          & 63.67          & 64.04          & 63.89 \\
w/o decay     & 60.66          & 59.01          & 59.31          & 64.78          & 65.07          & 63.78 \\
w/o both      & 59.31          & 58.12          & 58.79          & 63.67          & 64.04          & 63.89 \\
CREST (full)  & \textbf{60.21} & \textbf{59.88} & \textbf{59.16} & \textbf{65.04} & \textbf{65.04} & \textbf{64.19} \\
\bottomrule
\end{tabular}
\end{table}

\begin{figure*}[h]
    \centering
    \includegraphics[width=1\linewidth]{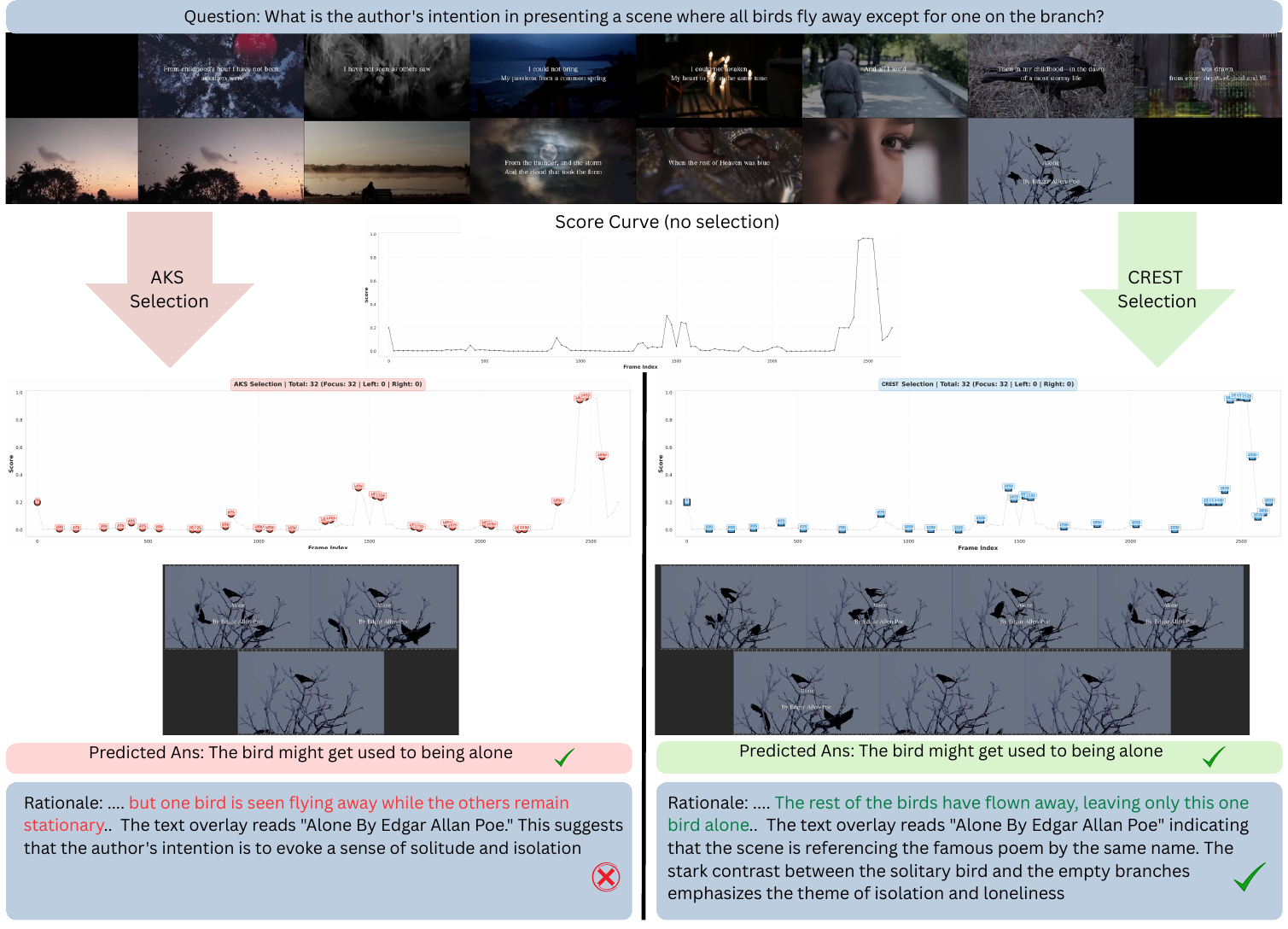}
    \caption{Both methods predict the correct answer, but differ in evidence quality. 
Rationales shown are truncated for brevity (indicated by ``...''); full descriptions 
are provided in Appendix~\ref{Elaborated_description}. \textsc{AKS} misses the 
critical temporal transition, producing an incomplete and partially inconsistent 
rationale. \textsc{CREST} captures the key event through curvature-aware dense 
sampling, yielding a coherent, visually grounded explanation. Correct answers alone 
do not guarantee faithful reasoning.}
    \label{fig:verifiability_event}
\end{figure*}
CREST outperforms AKS at every frame budget on both benchmarks
(Table~\ref{tab:budget}). The smaller marginal gain from $k{=}16$ to $k{=}32$ for
CREST compared to AKS suggests that CREST extracts more information from each selected
frame, rather than relying on increased budget to compensate for suboptimal selection. CREST is robust to the choice of vision-language scorer, outperforming AKS consistently
across BLIP, CLIP, and Sevila (Table~\ref{tab:vl_scorer}). This confirms that the gains
arise from the temporal selection mechanism itself, not from a dependency on any
particular relevance model.

Curvature is the critical component. Removing it collapses CREST to a uniform
suppression strategy and degrades performance across all scorers and both datasets
(Table~\ref{tab:ablation}), confirming that local geometric structure in the relevance
signal, not score magnitude, drives selection quality. Temporal decay has a smaller but
consistent effect: by progressively relaxing earlier suppression decisions, it allows
the budget to reach informative regions that would otherwise remain excluded. Removing
both components produces the largest drop, establishing that curvature modulation and
temporal decay contribute independently and complementarily.

% -----------------------------------------------------------------------
% FIGURES 4 & 5 — placed here, BEFORE \subsection{Grounding...}
%
% In two-column mode, figure* can only anchor to the TOP of a page [t].
% LaTeX scans forward from the declaration point to find the next free
% top slot. Declaring them here, just before the subsection, gives them
% the best chance of landing on the same page as the text that cites them.
%
% Rules for figure* placement in two-column:
%   DO     use [t]                     — the only reliable specifier
%   DO NOT use [h] or [H]             — silently ignored, causes drift
%   DO NOT use \FloatBarrier before   — closes the queue, pushes figures later
% -----------------------------------------------------------------------

\begin{figure*}[t]
    \centering
    \includegraphics[width=1\linewidth]{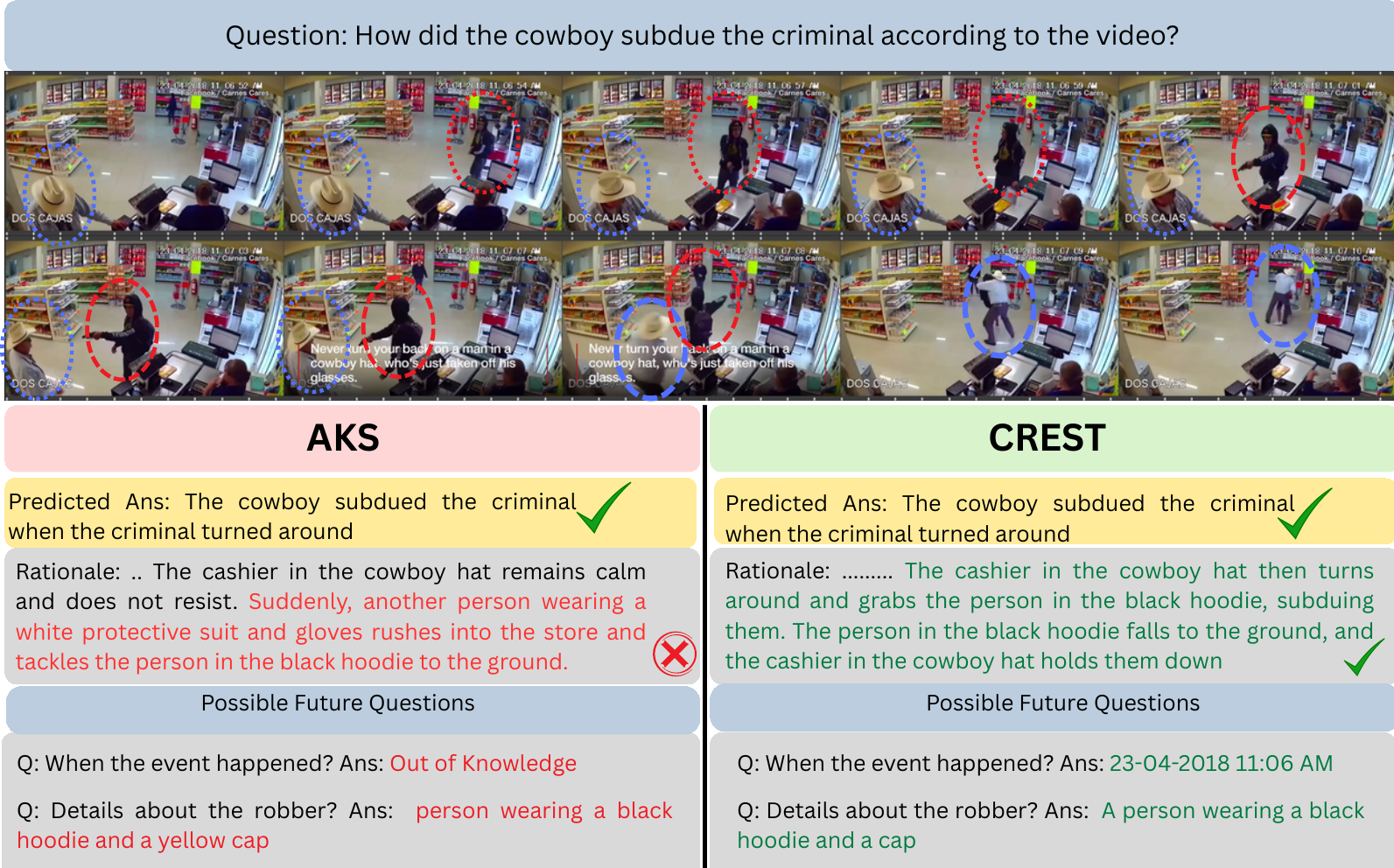}
    \caption{Insufficient grounding leads to hallucinated rationale and failure on 
follow-up queries. \textsc{CREST} preserves critical temporal and contextual 
information, supporting consistent downstream reasoning. Rationales shown are 
truncated for brevity (indicated by ``...''); full descriptions are provided in 
Appendix~\ref{Elaborated_description}.}
    \label{fig:future_qa}
\end{figure*}

\subsection{Grounding Quality Evaluation}
\label{subsec:grounding}
Multiple-choice accuracy does not fully measure whether predictions are grounded in
visual evidence: correct answers can arise from partial cues without genuine
comprehension. We therefore assess whether selected frames support coherent video
description, which requires capturing both salient events and sufficient temporal
context.

Figure~\ref{fig:verifiability_event} shows that even when both methods answer correctly,
their underlying evidence differs substantially. AKS misses the key temporal transition,
producing a fragmented rationale. CREST, by concentrating selections around
high-curvature regions, recovers the critical event and produces a coherent, visually
grounded explanation. Figure~\ref{fig:future_qa} further illustrates the downstream
consequence: insufficient grounding in AKS leads to hallucinated details and failure on follow-up queries, while CREST preserves contextual evidence that enables accurate responses. We quantify this using an LLM-as-a-judge protocol, in which descriptions generated from frames selected by each method are pairwise compared by a stronger model. We use two
independent judges and average results over three runs with randomized response order
(variation $<$0.5\%; ties excluded).

Table~\ref{tab:llm_judge_standard} reports the results, where CREST achieves higher win rates across both judges and both benchmarks. The high
inter-judge agreement (82.72\% on LVB, 79.52\% on VideoMME) indicates that the
preference for CREST-generated descriptions is consistent across models rather than an
artifact of a single judge's stylistic bias.

\begin{table}[t]
\centering
\footnotesize
\setlength{\tabcolsep}{5pt}
\caption{LLM-as-a-judge win rates (\%) on standard long-video benchmarks. Numbers show the percentage of questions where each method is preferred (ties excluded; 3-run average, variation ${<}0.5\%$). Judge agreement: 82.72\% (LVB), 79.52\% (V-MME).}
\label{tab:llm_judge_standard}
\begin{tabular}{l cccc}
\toprule
& \multicolumn{2}{c}{Gemini-3-Flash} & \multicolumn{2}{c}{GPT-5.4} \\
\cmidrule(lr){2-3}\cmidrule(lr){4-5}
Benchmark & CREST & AKS & CREST & AKS \\
\midrule
LongVideoBench & \textbf{60.58} & 39.42 & \textbf{59.46} & 40.54 \\
VideoMME       & \textbf{54.50} & 45.50 & \textbf{54.30} & 45.70 \\
\bottomrule
\end{tabular}
\end{table}

\subsection{TempRel: A Benchmark for Different Temporal Relevance Structures}
\label{subsec:temprel}
% Table 6: 3 cols — fits in one column

% To analyze the performance under different temporal relevance structures, we introduce a diagnostic benchmark, TempRel. TempRel evaluates frame selection under two controlled temporal regimes:
% Extended-Relevance (ER), where relevant information is distributed over a long temporal
% spans, and Hierarchical-Relevance (HR), where dominant peaks coexist with secondary informative events. CREST outperforms AKS in both settings
% (Table~\ref{tab:temprel_results}). The improvement is most pronounced in ER, where
% methods without modeling long-range temporal structure cluster near early
% peaks and miss later evidence. In the HR setting, CREST's curvature-modulated radius balances dominant and secondary peaks, confirming adaptability to heterogeneous relevance distributions.

To analyze performance under different temporal relevance structures, 
we introduce TempRel, a diagnostic benchmark evaluating frame selection 
under two controlled regimes: Extended-Relevance (ER), where relevant 
information spans long temporal windows, and Hierarchical-Relevance (HR), 
where dominant peaks coexist with secondary informative events. CREST 
outperforms AKS in both settings (Table~\ref{tab:temprel_results}). The 
improvement is most pronounced in ER, where methods lacking long-range 
temporal modeling cluster near early peaks and miss later evidence. In HR, 
CREST's curvature-modulated radius balances dominant and secondary peaks, 
confirming adaptability to heterogeneous relevance distributions.

\begin{table}[h]
\centering
\small
\caption{VQA accuracy (\%) on TempRel under different relevance regimes. All methods
use the same backbone, scorer, and frame budget.}
\label{tab:temprel_results}
\begin{tabular}{lccc}
\toprule
Method & ER & HR & Overall \\
\midrule
AKS & 62.22 & 76.16 & 68.96 \\
CREST (ours) & \textbf{67.13} & \textbf{80.83} & \textbf{73.64} \\
\bottomrule
\end{tabular}
\end{table}

For grounding quality, CREST outperforms AKS in both the context
and extended regimes (Table~\ref{tab:llm_judge_temprel}). 
% The advantage is larger in the extended setting, consistent with
% the accuracy results in Table~\ref{tab:llm_judge_standard}, and confirms that curvature-aware selection produces better-grounded descriptions when relevant information is distributed over longer spans.

\subsection{Statistical significance}
The VQA accuracy gains over AKS are consistent across benchmarks, frame budgets, and relevance scorers, but modest in magnitude. Stronger evidence comes from the LLM-as-a-judge evaluation: a binomial test rejects no preference ($H_0: p = 0.50$) at $\alpha = 0.01$ across both judges and all benchmarks, indicating that CREST selects frames that support more coherent, visually grounded explanations even when both methods reach the same final answer. These results suggest that accuracy alone is an insufficient measure of frame selection quality in long-video understanding.

\begin{table}[t]
\centering
\footnotesize
\setlength{\tabcolsep}{5pt}
\caption{LLM-as-a-judge win rates (\%) on TempRel under different temporal relevance structures. Numbers show the percentage of questions where each method is preferred (ties excluded; 3-run average, variation ${<}0.5\%$). Sample counts: Context 193, Extended 355, Total 548.}
\label{tab:llm_judge_temprel}
\begin{tabular}{l cccc}
\toprule
& \multicolumn{2}{c}{Gemini-3-Flash} & \multicolumn{2}{c}{GPT-5.4} \\
\cmidrule(lr){2-3}\cmidrule(lr){4-5}
Setting & CREST & AKS & CREST & AKS \\
\midrule
Context  & \textbf{53.89} & 46.11 & \textbf{54.92} & 45.08 \\
Extended & \textbf{57.75} & 42.25 & \textbf{55.77} & 44.23 \\
Total    & \textbf{56.39} & 43.61 & \textbf{55.47} & 44.53 \\
\bottomrule
\end{tabular}
\end{table}

\section{Conclusions}
\label{sec:conclusion}

% We introduce CREST, a training-free frame selection method for efficient long-video understanding. CREST treats query–frame relevance as a temporal signal and uses local curvature to allocate a limited frame budget more effectively across brief decisive events and slowly evolving evidence. Experiments on LongVideoBench and VideoMME show that CREST gives consistent improvements over lightweight baselines under matched backbone and frame-budget settings, while remaining competitive with stronger retrieval-based methods such as MIRA at much lower preprocessing cost. Through TempRel, we further evaluate CREST under controlled temporal relevance regimes and show that it handles both extended and hierarchical evidence structures. Beyond answer accuracy, pairwise LLM-as-a-judge evaluation indicates that CREST-selected frames support more informative and visually grounded descriptions. Overall, these results suggest that modeling the temporal geometry of relevance provides a simple, scalable, and practical direction for long-video understanding under constrained computational budgets.

We introduce CREST, a training-free frame selection method for efficient long-video understanding. CREST treats query--frame relevance as a temporal signal and uses local curvature to allocate a limited frame budget more effectively across brief decisive events and slowly evolving evidence. Experiments on LongVideoBench and VideoMME show that CREST gives consistent improvements over lightweight baselines under matched backbone and frame-budget settings, while remaining competitive with stronger retrieval-based methods such as MIRA at much lower preprocessing cost. Through TempRel, we further show that CREST handles both extended and hierarchical evidence structures. Beyond answer accuracy, pairwise LLM-as-a-judge evaluation indicates that CREST-selected frames support more informative and visually grounded descriptions. Overall, these results suggest that modeling the temporal geometry of relevance provides a simple, scalable, and practical direction for long-video understanding under constrained computational budgets.

%We introduced CREST, a curvature-aware and training-free frame selection method that explicitly models the temporal structure of relevance for long-video understanding. By leveraging local geometric properties of the relevance signal, CREST enables adaptive and
% content-aware allocation of a limited frame budget, effectively balancing informativeness
% and redundancy without requiring additional training or multi-stage retrieval pipelines.
% Extensive experiments on LongVideoBench and VideoMME show that CREST consistently
% outperforms prior lightweight methods under identical settings, while remaining
% competitive with more complex approaches such as MIRA at a small fraction of the
% computational cost. Through the proposed TempRel diagnostic dataset, we further
% demonstrate that CREST generalizes across diverse temporal relevance regimes, effectively
% handling both extended and hierarchical structures. Beyond answer accuracy, LLM-as-a-judge
% evaluation indicates that CREST produces more informative and better-grounded video
% descriptions, reflecting improved alignment between selected frames and visual evidence.
% Overall, the results show that explicitly modeling temporal relevance provides a simple
% yet effective and scalable foundation for efficient long-video understanding under
% constrained computational budgets.

\section*{Limitations}

The accuracy comparison in Table~\ref{tab:overall_comparison} is conducted under
different frame budgets: CREST passes $M = 32$ frames to the downstream MLLM,
while MIRA uses $M = 64$. Computational constraints prevented evaluation at
$M = 64$. The reported accuracy gap therefore reflects a combination of selection
quality and budget difference; disentangling the two requires running CREST at
$M = 64$, which we leave to future work.

The decay parameter $\rho$ was fixed at $0.5$ across all experiments without
tuning. This value produced consistent results on the benchmarks evaluated, but
the sensitivity of CREST to $\rho$ across domains with different temporal
structures remains uncharacterized. Domain-specific tuning may further improve
performance in settings where relevant content is distributed differently from
standard benchmarks.

CREST is query-conditioned by design, requiring the natural-language query to be
available at frame selection time. This is well-suited to the video question
answering setting studied here, but rules out offline use cases such as
query-agnostic video summarization or pre-computed frame indices.

\bibliography{custom}

@article{liu2023llava,
  title        = {Visual Instruction Tuning},
  author       = {Liu, Haotian and Li, Chunyuan and Wu, Qingyang and Lee, Yong Jae},
  journal      = {arXiv preprint arXiv:2304.08485},
  year         = {2023},
  eprint       = {2304.08485},
  archivePrefix= {arXiv},
  primaryClass = {cs.CV},
  url          = {https://arxiv.org/abs/2304.08485}
}

@article{bai2023qwenvl,
  title        = {Qwen-{VL}: A Versatile Vision-Language Model for Understanding,
                  Localization, Text Reading, and Beyond},
  author       = {Bai, Jinze and Bai, Shuai and Yang, Shusheng and Wang, Shijie
                  and Tan, Sinan and Wang, Peng and Lin, Junyang and Zhou, Chang
                  and Zhou, Jingren},
  journal      = {arXiv preprint arXiv:2308.12966},
  year         = {2023},
  eprint       = {2308.12966},
  archivePrefix= {arXiv},
  primaryClass = {cs.CV},
  url          = {https://arxiv.org/abs/2308.12966}
}

@article{zhu2023minigpt4,
  title        = {{MiniGPT-4}: Enhancing Vision-Language Understanding with
                  Advanced Large Language Models},
  author       = {Zhu, Deyao and Chen, Jun and Shen, Xiaohui and Li, Xiang and
                  Elhoseiny, Mohamed},
  journal      = {arXiv preprint arXiv:2304.10592},
  year         = {2023},
  eprint       = {2304.10592},
  archivePrefix= {arXiv},
  primaryClass = {cs.CV},
  url          = {https://arxiv.org/abs/2304.10592}
}

@article{chen2023internvl,
  title        = {{InternVL}: Scaling up Vision Foundation Models and Aligning
                  for Generic Visual-Linguistic Tasks},
  author       = {Chen, Zhe and Wu, Jiannan and Wang, Wenhai and Su, Weijie and
                  Chen, Guo and Xing, Sen and Zhong, Muyan and Zhang, Qinglong and
                  Zhu, Xizhou and Lu, Lewei and Li, Bin and Luo, Ping and
                  Lu, Tong and Qiao, Yu and Dai, Jifeng},
  journal      = {arXiv preprint arXiv:2312.14238},
  year         = {2023},
  eprint       = {2312.14238},
  archivePrefix= {arXiv},
  primaryClass = {cs.CV},
  url          = {https://arxiv.org/abs/2312.14238}
}

@article{zhang2024llavavideo,
  title        = {{LLaVA-Video}: Video Instruction Tuning With Synthetic Data},
  author       = {Zhang, Yuanhan and Wu, Jinming and Li, Wei and Li, Bo and
                  Ma, Zejun and Liu, Ziwei and Li, Chunyuan},
  journal      = {arXiv preprint arXiv:2410.02713},
  year         = {2024},
  eprint       = {2410.02713},
  archivePrefix= {arXiv},
  primaryClass = {cs.CV},
  url          = {https://arxiv.org/abs/2410.02713}
}

@article{lin2023videollava,
  title        = {{Video-LLaVA}: Learning United Visual Representation by
                  Alignment Before Projection},
  author       = {Lin, Bin and Zhu, Bin and Ye, Yang and Ning, Munan and
                  Jin, Peng and Yuan, Li},
  journal      = {arXiv preprint arXiv:2311.10122},
  year         = {2023},
  eprint       = {2311.10122},
  archivePrefix= {arXiv},
  primaryClass = {cs.CV},
  url          = {https://arxiv.org/abs/2311.10122}
}

@article{xu2024pllava,
  title        = {{PLLaVA}: Parameter-Free {LLaVA} Extension from Images to
                  Videos for Video Dense Captioning},
  author       = {Xu, Lin and Zhao, Yilin and Zhou, Daquan and Lin, Zhijie and
                  Ng, See Kiong and Feng, Jiashi},
  journal      = {arXiv preprint arXiv:2404.16994},
  year         = {2024},
  eprint       = {2404.16994},
  archivePrefix= {arXiv},
  primaryClass = {cs.CV},
  url          = {https://arxiv.org/abs/2404.16994}
}

@article{xu2024sfllava,
  title        = {{SlowFast-LLaVA}: A Strong Training-Free Baseline for Video
                  Large Language Models},
  author       = {Xu, Mingze and Gao, Mingfei and Gan, Zhe and Chen, Hong-You and
                  Lai, Zhengfeng and Gang, Haiming and Kang, Kai and Dehghan, Afshin},
  journal      = {arXiv preprint arXiv:2407.15841},
  year         = {2024},
  eprint       = {2407.15841},
  archivePrefix= {arXiv},
  primaryClass = {cs.CV},
  url          = {https://arxiv.org/abs/2407.15841}
}

@article{tang2025aks,
  title        = {Adaptive Keyframe Sampling for Long Video Understanding},
  author       = {Tang, Xi and Qiu, Jihao and Xie, Lingxi and Tian, Yunjie and
                  Jiao, Jianbin and Ye, Qixiang},
  journal      = {arXiv preprint arXiv:2502.21271},
  year         = {2025},
  eprint       = {2502.21271},
  archivePrefix= {arXiv},
  primaryClass = {cs.CV},
  url          = {https://arxiv.org/abs/2502.21271}
}

@article{hao2026mira,
  title        = {{MIRA}: Multi-view Information Retrieval with Adaptive Routing
                  for Test-time Long-video Comprehension},
  author       = {Hao, Zecheng and Ma, Wenxuan and Cui, Yufeng and Li, Shuang and
                  Wang, Xinlong and Huang, Tiejun},
  journal      = {Transactions on Machine Learning Research},
  year         = {2026}
}

@inproceedings{vilamp2025,
  title        = {Scaling Video-Language Models to 10K Frames via Hierarchical
                  Differential Distillation},
  author       = {Cheng, Chuanqi and Guan, Jian and Wu, Wei and Yan, Rui},
  booktitle    = {Proceedings of the International Conference on Machine Learning
                  (ICML)},
  year         = {2025},
  eprint       = {2504.02438},
  archivePrefix= {arXiv},
  primaryClass = {cs.CV},
  url          = {https://arxiv.org/abs/2504.02438}
}

@article{bimba2025,
  title        = {{BIMBA}: Selective-Scan Compression for Long-Range Video
                  Question Answering},
  author       = {Islam, Md Mohaiminul and Nagarajan, Tushar and Wang, Huiyu and
                  Bertasius, Gedas and Torresani, Lorenzo},
  journal      = {arXiv preprint arXiv:2503.09590},
  year         = {2025},
  eprint       = {2503.09590},
  archivePrefix= {arXiv},
  primaryClass = {cs.CV},
  url          = {https://arxiv.org/abs/2503.09590}
}

@article{videotree2025,
  title        = {{VideoTree}: Adaptive Tree-based Video Representation for
                  {LLM} Reasoning on Long Videos},
  author       = {Wang, Ziyang and Yu, Shoubin and Stengel-Eskin, Elias and
                  Yoon, Jaehong and Cheng, Feng and Bertasius, Gedas and
                  Bansal, Mohit},
  journal      = {arXiv preprint arXiv:2405.19209},
  year         = {2024},
  eprint       = {2405.19209},
  archivePrefix= {arXiv},
  primaryClass = {cs.CV},
  url          = {https://arxiv.org/abs/2405.19209}
}

@inproceedings{bolt2025,
  title        = {{BOLT}: Boost Large Vision-Language Model Without Training
                  for Long-form Video Understanding},
  author       = {Liu, Shuming and Zhao, Chen and Xu, Tianqi and Ghanem, Bernard},
  booktitle    = {Proceedings of the IEEE/CVF Conference on Computer Vision and
                  Pattern Recognition (CVPR)},
  year         = {2025},
  eprint       = {2503.21483},
  archivePrefix= {arXiv},
  primaryClass = {cs.CV},
  url          = {https://arxiv.org/abs/2503.21483}
}

@inproceedings{yuan2025dtoma,
  title        = {Training-free Dynamic Token {MA}nipulation for Long Video
                  Understanding},
  author       = {Yuan, Rui and others},
  booktitle    = {Proceedings of the Thirty-Fourth International Joint Conference
                  on Artificial Intelligence (IJCAI)},
  pages        = {2314--2322},
  year         = {2025},
  doi          = {10.24963/ijcai.2025/258},
  url          = {https://www.ijcai.org/proceedings/2025/258}
}

@inproceedings{liu2024vila,
  title        = {{VILA}: On Pre-training for Visual Language Models},
  author       = {Lin, Ji and Yin, Hongxu and Ping, Wei and Lu, Yao and
                  Molchanov, Pavlo and Tao, Andrew and Mao, Huizi and
                  Kautz, Jan and Shoeybi, Mohammad and Han, Song},
  booktitle    = {Proceedings of the IEEE/CVF Conference on Computer Vision and
                  Pattern Recognition (CVPR)},
  year         = {2024},
  eprint       = {2312.07533},
  archivePrefix= {arXiv},
  primaryClass = {cs.CV},
  url          = {https://arxiv.org/abs/2312.07533}
}

@article{minicpmv2024,
  title        = {{MiniCPM-V}: A {GPT-4V} Level {MLLM} on Your Phone},
  author       = {Yao, Yuan and Yu, Tianyu and Zhang, Ao and Wang, Chongyi and
                  Cui, Junbo and Zhu, Hongji and Cai, Tianchi and Li, Haoyu and
                  Zhao, Weilin and He, Zhihui and Chen, Qianyu and Zhou, Huarong and
                  Zou, Zhensheng and Zhang, Haoye and Hu, Shengding and Zheng, Zhi
                  and Zhou, Jie and Cai, Jie and Han, Xu and Zeng, Guoyang and
                  Li, Dahai and Liu, Zhiyuan and Sun, Maosong},
  journal      = {arXiv preprint arXiv:2408.01800},
  year         = {2024},
  eprint       = {2408.01800},
  archivePrefix= {arXiv},
  primaryClass = {cs.CV},
  url          = {https://arxiv.org/abs/2408.01800}
}

@misc{openai2023gpt4v,
  title        = {{GPT-4V}(ision) System Card},
  author       = {{OpenAI}},
  year         = {2023},
  howpublished = {\url{https://openai.com/research/gpt-4v-system-card}}
}

@misc{openai2024gpt4o,
  title        = {{GPT-4o} System Card},
  author       = {{OpenAI}},
  year         = {2024},
  howpublished = {\url{https://openai.com/research/gpt-4o-system-card}}
}

@misc{gemini15,
  title        = {Gemini 1.5: Unlocking Multimodal Understanding across Millions
                  of Tokens of Context},
  author       = {{Google DeepMind}},
  year         = {2024},
  eprint       = {2403.05530},
  archivePrefix= {arXiv},
  primaryClass = {cs.CL},
  url          = {https://arxiv.org/abs/2403.05530}
}

@inproceedings{liu2025nvila,
  title        = {{NVILA}: Efficient Frontier Visual Language Models},
  author       = {Liu, Zhijian and Zhu, Ligeng and Shi, Baifeng and
                  Zhang, Zhuoyang and Lou, Yuming and Yang, Shang and
                  Xi, Haocheng and Cao, Shiyi and Gu, Yuxian and Li, Dacheng and
                  Li, Xiuyu and Fang, Yunhao and Chen, Yukang and Hsieh, Cheng-Yu
                  and Huang, De-An and Cheng, An-Chieh and Nath, Vishwesh and
                  Hu, Jinyi and Liu, Sifei and Krishna, Ranjay and Xu, Daguang and
                  Wang, Xiaolong and Molchanov, Pavlo and Kautz, Jan and
                  Yin, Hongxu and Han, Song and Lu, Yao},
  booktitle    = {Proceedings of the IEEE/CVF Conference on Computer Vision and
                  Pattern Recognition (CVPR)},
  year         = {2025},
  eprint       = {2412.04468},
  archivePrefix= {arXiv},
  primaryClass = {cs.CV},
  url          = {https://arxiv.org/abs/2412.04468}
}

@inproceedings{zohar2025apollo,
  title        = {Apollo: An Exploration of Video Understanding in Large
                  Multimodal Models},
  author       = {Zohar, Orr and Wang, Xiaohan and Dubois, Yann and Mehta, Nikhil
                  and Xiao, Tong and Hansen-Estruch, Philippe and Yu, Licheng and
                  Wang, Xiaofang and Juefei-Xu, Felix and Zhang, Ning and
                  Yeung-Levy, Serena and Xia, Xide},
  booktitle    = {Proceedings of the IEEE/CVF Conference on Computer Vision and
                  Pattern Recognition (CVPR)},
  year         = {2025},
  eprint       = {2412.10360},
  archivePrefix= {arXiv},
  primaryClass = {cs.CV},
  url          = {https://arxiv.org/abs/2412.10360}
}

@article{wu2024longvideobench,
  title        = {{LongVideoBench}: A Benchmark for Long-context Interleaved
                  Video-Language Understanding},
  author       = {Wu, Haoning and Li, Dongxu and Chen, Bei and Li, Junnan},
  journal      = {arXiv preprint arXiv:2407.15754},
  year         = {2024},
  eprint       = {2407.15754},
  archivePrefix= {arXiv},
  primaryClass = {cs.CV},
  url          = {https://arxiv.org/abs/2407.15754}
}

@article{fu2024videomme,
  title        = {{Video-MME}: The First-Ever Comprehensive Evaluation Benchmark
                  of Multi-modal {LLMs} in Video Analysis},
  author       = {Fu, Chaoyou and Dai, Yuhan and Luo, Yongdong and Li, Lei and
                  Ren, Shuhuai and Zhang, Renrui and Wang, Zihan and Zhou, Chenyu
                  and Shen, Yunhang and Zhang, Mengdan and Chen, Peixian and
                  Li, Yanwei and Lin, Shaohui and Zhao, Sirui and Li, Ke and
                  Xu, Tong and Zheng, Xiawu and Chen, Enhong and Ji, Rongrong and
                  Sun, Xing},
  journal      = {arXiv preprint arXiv:2405.21075},
  year         = {2024},
  eprint       = {2405.21075},
  archivePrefix= {arXiv},
  primaryClass = {cs.CV},
  url          = {https://arxiv.org/abs/2405.21075}
}

@inproceedings{zhang2025lmmseval,
  title        = {{LMMs-Eval}: Reality Check on the Evaluation of Large
                  Multimodal Models},
  author       = {Zhang, Kaichen and Li, Bo and Zhang, Peiyuan and Pu, Fanyi and
                  Cahyono, Joshua Adrian and Hu, Kairui and Liu, Shuai and
                  Zhang, Yuanhan and Yang, Jingkang and Li, Chunyuan and
                  Liu, Ziwei},
  booktitle    = {Findings of the Association for Computational Linguistics:
                  NAACL 2025},
  pages        = {881--916},
  year         = {2025},
  url          = {https://aclanthology.org/2025.findings-naacl.51/}
}

@article{tstar2025,
  title        = {{T*}: Re-thinking Temporal Search for Long-Form Video
                  Understanding},
  author       = {Ye, Jinhui and Wang, Zihan and Sun, Haosen and
                  Chandrasegaran, Keshigeyan and Durante, Zane and
                  Eyzaguirre, Cristobal and Bisk, Yonatan and
                  Niebles, Juan Carlos and Adeli, Ehsan and Fei-Fei, Li and
                  Wu, Jiajun and Li, Manling},
  journal      = {arXiv preprint arXiv:2504.02259},
  year         = {2025},
  eprint       = {2504.02259},
  archivePrefix= {arXiv},
  primaryClass = {cs.CV},
  url          = {https://arxiv.org/abs/2504.02259}
}

@article{framevoyager2024,
  title        = {Frame-Voyager: Learning to Query Frames for Video Large
                  Language Models},
  author       = {Yu, Sicheng and Jin, Chengkai and Wang, Huanyu and Chen, Zhenghao
                  and Jin, Sheng and Zuo, Zhongrong and Xu, Xiaolei and
                  Sun, Zhenbang and Zhang, Bingni and Wu, Jiawei and Zhang, Hao
                  and Sun, Qianru},
  journal      = {arXiv preprint arXiv:2410.03226},
  year         = {2024},
  eprint       = {2410.03226},
  archivePrefix= {arXiv},
  primaryClass = {cs.CV},
  url          = {https://arxiv.org/abs/2410.03226}
}

@article{tang2025tspo,
  title        = {{TSPO}: Temporal Sampling Policy Optimization for Long-form
                  Video Language Understanding},
  author       = {Tang, Canhui and Han, Zifan and Sun, Hongbo and Zhou, Sanping
                  and Zhang, Xuchong and Wei, Xin and Yuan, Ye and Zhang, Huayu
                  and Xu, Jinglin and Sun, Hao},
  journal      = {arXiv preprint arXiv:2508.04369},
  year         = {2025},
  eprint       = {2508.04369},
  archivePrefix= {arXiv},
  primaryClass = {cs.CV},
  note         = {Accepted at AAAI 2026},
  url          = {https://arxiv.org/abs/2508.04369}
}

@article{li2025tpo,
  title        = {Temporal Preference Optimization for Long-Form Video
                  Understanding},
  author       = {Li, Rui and Wang, Xiaohan and Zhang, Yuhui and Wang, Zeyu and
                  Yeung-Levy, Serena},
  journal      = {arXiv preprint arXiv:2501.13919},
  year         = {2025},
  eprint       = {2501.13919},
  archivePrefix= {arXiv},
  primaryClass = {cs.CV},
  url          = {https://arxiv.org/abs/2501.13919}
}

@article{wang2025adaretake,
  title        = {{AdaReTaKe}: Adaptive Redundancy Reduction to Perceive Longer
                  for Video-language Understanding},
  author       = {Wang, Xiao and Si, Qingyi and Wu, Jianlong and Zhu, Shiyu and
                  Cao, Li and Nie, Liqiang},
  journal      = {arXiv preprint arXiv:2503.12559},
  year         = {2025},
  eprint       = {2503.12559},
  archivePrefix= {arXiv},
  primaryClass = {cs.CV},
  url          = {https://arxiv.org/abs/2503.12559}
}

@inproceedings{hu2025dual,
  title        = {{M-LLM} Based Video Frame Selection for Efficient Video
                  Understanding},
  author       = {Hu, Kai and Gao, Feng and Nie, Xiaohan and Zhou, Peng and
                  Tran, Son and Neiman, Tal and Wang, Lingyun and Shah, Mubarak and
                  Hamid, Raffay and Yin, Bing and Chilimbi, Trishul},
  booktitle    = {Proceedings of the IEEE/CVF Conference on Computer Vision and
                  Pattern Recognition (CVPR)},
  year         = {2025},
  eprint       = {2502.19680},
  archivePrefix= {arXiv},
  primaryClass = {cs.CV},
  url          = {https://arxiv.org/abs/2502.19680}
}

@article{hu2025cos,
  title        = {{CoS}: Chain-of-Shot Prompting for Long Video Understanding},
  author       = {Hu, Jian and Cheng, Zixu and Si, Chenyang and Li, Wei and
                  Gong, Shaogang},
  journal      = {arXiv preprint arXiv:2502.06428},
  year         = {2025},
  eprint       = {2502.06428},
  archivePrefix= {arXiv},
  primaryClass = {cs.CV},
  url          = {https://arxiv.org/abs/2502.06428}
}

@article{fang2025narkfc,
  title        = {Threading Keyframe with Narratives: {MLLMs} as Strong Long
                  Video Comprehenders},
  author       = {Fang, Bo and Wu, Wenhao and Wu, Qiangqiang and Song, Yuxin
                  and Chan, Antoni B.},
  journal      = {arXiv preprint arXiv:2505.24158},
  year         = {2025},
  eprint       = {2505.24158},
  archivePrefix= {arXiv},
  primaryClass = {cs.CV},
  url          = {https://arxiv.org/abs/2505.24158}
}

@article{gao2025apvr,
  title        = {{APVR}: Hour-Level Long Video Understanding with Adaptive Pivot
                  Visual Information Retrieval},
  author       = {Gao, Hong and Bao, Yiming and Tu, Xuezhen and Zhong, Bin and
                  Yue, Linan and Zhang, Minling},
  journal      = {arXiv preprint arXiv:2506.04953},
  year         = {2025},
  eprint       = {2506.04953},
  archivePrefix= {arXiv},
  primaryClass = {cs.CV},
  url          = {https://arxiv.org/abs/2506.04953}
}

@inproceedings{wang2024videoagent,
  title        = {{VideoAgent}: Long-Form Video Understanding with Large Language
                  Model as Agent},
  author       = {Wang, Xiaohan and Zhang, Yuhui and Zohar, Orr and
                  Yeung-Levy, Serena},
  booktitle    = {Proceedings of the European Conference on Computer Vision
                  (ECCV)},
  year         = {2024},
  eprint       = {2403.10517},
  archivePrefix= {arXiv},
  primaryClass = {cs.CV},
  url          = {https://arxiv.org/abs/2403.10517}
}

@article{xu2025evrag,
  title        = {{E-VRAG}: Enhancing Long Video Understanding with
                  Resource-Efficient Retrieval Augmented Generation},
  author       = {Xu, Zeyu and Zhang, Junkang and Wang, Qiang and Liu, Yi},
  journal      = {arXiv preprint arXiv:2508.01546},
  year         = {2025},
  eprint       = {2508.01546},
  archivePrefix= {arXiv},
  primaryClass = {cs.CV},
  url          = {https://arxiv.org/abs/2508.01546}
}

@article{chen2024longvila,
  title        = {{LongVILA}: Scaling Long-Context Visual Language Models for
                  Long Videos},
  author       = {Chen, Yukang and Xue, Fuzhao and Li, Dacheng and Hu, Qinghao
                  and Zhu, Ligeng and Li, Xiuyu and Fang, Yunhao and Tang, Haotian
                  and Yang, Shang and Liu, Zhijian and Han, Song and Molchanov,
                  Pavlo and Kautz, Jan},
  journal      = {arXiv preprint arXiv:2408.10188},
  year         = {2024},
  eprint       = {2408.10188},
  archivePrefix= {arXiv},
  primaryClass = {cs.CV},
  url          = {https://arxiv.org/abs/2408.10188}
}

@inproceedings{shen2024longvu,
  title        = {{LongVU}: Spatiotemporal Adaptive Compression for Long
                  Video-Language Understanding},
  author       = {Shen, Xiaoqian and Xiong, Yunyang and Zhao, Changsheng and
                  Wu, Lemeng and Chen, Jun and Zhu, Chenchen and Liu, Zechun and
                  Xiao, Fanyi and Varadarajan, Balakrishnan and Bordes, Florian and
                  Liu, Zhuang and Xu, Hu and Kim, Hyunwoo J. and Soran, Bilge and
                  Krishnamoorthi, Raghuraman and Elhoseiny, Mohamed and
                  Chandra, Vikas},
  booktitle    = {Proceedings of the International Conference on Machine Learning
                  (ICML)},
  year         = {2025},
  eprint       = {2410.17434},
  archivePrefix= {arXiv},
  primaryClass = {cs.CV},
  url          = {https://arxiv.org/abs/2410.17434}
}

\appendix

\section{Related Works}

Early MLLMs such as LLaVA~\citep{liu2023llava}, Qwen-VL~\citep{bai2023qwenvl},
MiniGPT-4~\citep{zhu2023minigpt4}, InternVL~\citep{chen2023internvl},
GPT-4o~\citep{openai2024gpt4o}, and NVILA~\citep{liu2025nvila} extend language models
to the visual domain by encoding images as visual tokens. Adapting these architectures
to video is non-trivial: Video-LLaVA~\citep{lin2023videollava} employs sparse uniform
sampling, which misses temporally localized events, while LLaVA-Video~%
\citep{zhang2024llavavideo} increases density but remains
constrained by context length. PLLaVA~\citep{xu2024pllava} improves parameter
efficiency, yet its performance is directly bounded by the quality of the frames it
receives. Across all these methods, frame selection is the binding constraint on
long-video reasoning.

One line of work learns frame importance through supervision. Frame-Voyager~%
\citep{framevoyager2024}, ViLaMP~\citep{vilamp2025}, and BIMBA~\citep{bimba2025}
incorporate learned scoring or redundancy reduction, while related efforts train
lightweight selectors from pseudo spatial and temporal labels~\citep{hu2025dual} or
combine supervised fine-tuning with preference optimization~\citep{li2025tpo}.
Apollo~\citep{zohar2025apollo} distills effective practices for long-range temporal
understanding, and TSPO~\citep{tang2025tspo} explores lightweight policy adaptation.
These methods improve accuracy but require additional data, training, and careful tuning,
limiting their generality across deployment settings.

Training-free methods avoid these costs by operating entirely at inference time.
Preprocessing approaches select frames before reasoning: VideoTree~\citep{videotree2025}
performs hierarchical search via clustering, CoS~\citep{hu2025cos} uses compact video
coding as a grounding proxy, and BOLT~\citep{bolt2025} relies on proxy relevance signals
from pretrained encoders. Retrieval-oriented methods perform multi-round temporal search
with object-centric cues~\citep{tstar2025} or interleave image-text streams
to preserve temporal continuity~\citep{fang2025narkfc}. Compression-based methods such
as SF-LLaVA~\citep{xu2024sfllava}, AdaRETAKE~\citep{wang2025adaretake}, and
APVR~\citep{gao2025apvr} reduce token counts through adaptive selection. Iterative
approaches refine the selected context during inference: VideoAgent~%
\citep{wang2024videoagent} queries frames by model confidence, T*~\citep{tstar2025}
uses multi-stage refinement, and E-VRAG~\citep{xu2025evrag} combines self-reflection
with hierarchical filtering. Despite their diversity, none of these methods explicitly
model the local temporal geometry of query-conditioned relevance.

The closest prior work to CREST imposes explicit structure on the selection process.
AKS~\citep{tang2025aks} formulates selection as a relevance--coverage trade-off, using
a recursive judge-and-split strategy to allocate frames across the video. Its behavior
is sensitive to hyperparameter choices and degrades toward near-uniform sampling when
relevance signals are diffuse. MIRA~\citep{hao2026mira} addresses expressiveness through
multi-view relevance scoring and adaptive routing, achieving strong accuracy at the cost
of substantial preprocessing overhead. AKS trades expressiveness for simplicity; MIRA
trades efficiency for accuracy. CREST occupies a different point in this space: it
requires no training, no multi-stage pipeline, and no dataset-specific tuning, while
directly modeling the curvature of the temporal relevance signal to enable adaptive,
event-centric frame selection.

\section{Appendix}

\subsection{Description Generation Prompt}
\label{Description_generation_prompt}

To assess whether selected frames support faithful and comprehensive video understanding, 
we generate structured descriptions using a standardized prompt. The prompt is designed 
to separate (i) question-specific visual evidence and (ii) broader domain coverage, 
ensuring that the generated descriptions capture both immediate relevance and contextual 
completeness.

The exact prompt used for description generation is as follows:

\begin{quote}
\textbf{You are an expert visual anthropologist analyzing a video clip frame-by-frame.}

Below is a multiple-choice question and its options.

\textbf{YOUR TASK IS NOT TO CHOOSE A, B, C, or D.}

Instead, generate a detailed, standalone two-part description of what the question is 
really asking about, based ONLY on what you perceive in the provided video frames.

\textbf{Part 1 -- Direct Frame Evidence (question-specific)} \\
Describe exactly what you see in the frames that relates directly to the question: 
objects, actions, sequence of events, timeline, text overlays, labels, diagrams, 
animations, people, environments, tools, or any other visual elements. Be exhaustive 
--- list every relevant visual detail and how they connect to the exact topic of the 
question. Imagine you are writing precise field notes that another researcher could use 
to reconstruct the exact scene without watching the video.

\textbf{Part 2 -- Full Domain Coverage (everything related that might be questioned 
later)} \\
Now describe everything related to the question's domain or topic that you can see 
anywhere in the frames. Cover every single aspect, sub-topic, variation, step, 
component, related object, concept, sequence, or idea that belongs to the same broader 
domain --- even if it is not directly mentioned in the question. Be completely exhaustive 
so that any future question on any part of this domain can be answered from this 
description alone. Include all key principles, methods, alternatives, applications, or 
connections that are visible in the video. Write as if you are creating a detailed 
encyclopedia entry that fully maps the entire topic shown in the video.

\textbf{Rules:}
\begin{itemize}
    \item Base everything strictly on visuals you actually see --- never invent or 
    assume content.
    \item Use precise, descriptive, and educational language.
    \item Do NOT mention the options, the correct answer, or say ``the question is 
    testing\ldots''.
    \item Do NOT summarize; be as detailed and comprehensive as possible in both parts.
    \item Structure your response clearly with Part 1 and Part 2 headings.
\end{itemize}

\textbf{Question:}
\end{quote}

\subsection{LLM-as-a-Judge Evaluation Prompt}
\label{LLM_as_a_judge_prompt}

To compare the quality of generated descriptions, we adopt an LLM-as-a-judge protocol. 
The evaluation focuses on two criteria with equal weight: (i) question-specific visual 
grounding and (ii) coverage of additional relevant context for future queries. The judge 
is instructed to make a strict binary decision between two descriptions, without allowing 
ties, and to focus exclusively on the defined evaluation criteria.

The exact prompt used for evaluation is as follows:

\begin{quote}
\textbf{You are a strict, impartial judge evaluating two already-generated descriptions 
from a video understanding benchmark.}

You will be given:
\begin{itemize}
    \item The original multiple-choice question
    \item The ground-truth correct answer (A, B, C, or D) --- used ONLY to define the 
    scene boundary
    \item Description 1 (Part 1 + Part 2 from Model 1)
    \item Description 2 (Part 1 + Part 2 from Model 2)
\end{itemize}

These descriptions were produced using a frame-grounded generation prompt. You MUST 
assume every statement in both descriptions is already strictly limited to visible 
frames. Your job is NOT to check grounding again.

\textbf{YOUR ONLY JOB:} Decide which description is BETTER overall, using exactly 
these two rubrics (equal weight):

\textbf{Rubric 1 -- Question-Specific Details (Part 1)} \\
Evaluate how precise, observable visual details are provided that directly relate to 
answering the given question. Score higher for:
\begin{itemize}
    \item More exhaustive listing of objects, actions, people, tools, text, labels, 
    spatial relationships, motion, and timeline elements that are relevant to the 
    question.
    \item Higher precision and reconstructability of the exact scene for this specific 
    question.
\end{itemize}

\textbf{Rubric 2 -- Related/Future-Question Coverage (Part 2)} \\
Evaluate how additional visible elements from the same scene/domain are included that 
could support future questions on the same topic. Score higher for:
\begin{itemize}
    \item More exhaustive coverage of other objects, steps, contextual elements, 
    variations, or supporting details inside the exact same scene (even if not needed 
    for the current question).
    \item Better readiness for follow-up questions without rewatching the video.
\end{itemize}

\textbf{Rules:}
\begin{itemize}
    \item Judge ONLY on the two rubrics above --- nothing else.
    \item Use the correct answer only to confirm the scene boundary.
    \item Ignore anything that is not related to the question at all.
    \item Be extremely critical: the description with clearly more details on BOTH 
    rubrics wins. No ties allowed.
    \item Penalize any detail that drifts outside the current question's scene.
\end{itemize}

\textbf{Response format (exactly like this):}

Comparison:

Rubric 1 analysis (question-specific details): [2--3 sentences comparing quantity and 
precision of details directly tied to the question]

Rubric 2 analysis (related/future-question coverage): [2--3 sentences comparing 
quantity and usefulness of additional scene elements for future questions]

Winner: [1 or 2]

\textbf{Now judge:}

Question: \{question\_text\}

Correct Answer: \{answer\_full\}

Description 1:

\{description\_1\}

Description 2:

\{description\_2\}

Judge now:
\end{quote}

\subsection{Qualitative Examples for Description Evaluation}
\label{Elaborated_description}

We provide representative qualitative examples comparing descriptions generated using 
frames selected by CREST and AKS. These examples illustrate differences in evidence 
grounding, temporal completeness, and susceptibility to hallucination.

\medskip
\noindent\textbf{Example 1: Convenience Store Robbery Scenario}

\noindent\textbf{Question:} How did the cowboy subdue the criminal according to the 
video? \\
\textbf{Correct Answer:} B. The cowboy subdued the criminal when the criminal turned 
around.

\medskip

\noindent\textbf{Description 1 (CREST)}

\textit{Part 1 -- Direct Frame Evidence (question-specific)} \\
The video shows a convenience store with two cashiers at the counter. The timestamp on 
the footage is ``23-04-2018 11:06 AM''. A person wearing a cowboy hat and a 
light-colored shirt is standing behind the counter, while another person in a dark blue 
shirt is working on the computer. A person wearing a black hoodie and a cap enters the 
store and approaches the counter. The person in the black hoodie pulls out a gun and 
points it at the cashier in the cowboy hat. The cashier in the cowboy hat remains calm 
and does not move. The person in the black hoodie demands money from the cashier. The 
cashier in the cowboy hat then turns around and grabs the person in the black hoodie, 
subduing them. The person in the black hoodie falls to the ground, and the cashier in 
the cowboy hat holds them down.

\textit{Part 2 -- Full Domain Coverage (everything related that might be questioned 
later)} \\
The video depicts a scenario where a person wearing a cowboy hat and a light-colored 
shirt subdues an armed robber in a convenience store. The cowboy uses physical strength 
and quick reflexes to overpower the criminal, who is armed with a gun. The scene 
highlights situational awareness, physical intervention, and the dynamics of an armed 
robbery scenario, including escalation, confrontation, and resolution. It also reflects 
elements of public safety, human response under threat, and interactions between 
civilians and offenders in a retail environment.

\medskip

\noindent\textbf{Description 2 (AKS)}

\textit{Part 1 -- Direct Frame Evidence (question-specific)} \\
The video shows a convenience store with shelves stocked with various products. There 
are two cashiers at the counter, one wearing a cowboy hat and the other in a blue shirt. 
A person wearing a black hoodie and a yellow cap enters the store. The person in the 
black hoodie points a gun at the cashier in the cowboy hat and demands money. The 
cashier in the cowboy hat remains calm and does not resist. Suddenly, another person 
wearing a white protective suit and gloves rushes into the store and tackles the person 
in the black hoodie to the ground. The person in the white suit restrains the person in 
the black hoodie until the police arrive.

\textit{Part 2 -- Full Domain Coverage (everything related that might be questioned 
later)} \\
The video depicts a robbery scenario in a convenience store involving an armed 
individual. The situation escalates when another individual intervenes and subdues the 
robber. The scene reflects themes of emergency response, intervention by third parties, 
and crime prevention in public environments. It emphasizes the role of external actors 
in mitigating threats and highlights the dynamics of unexpected intervention during 
criminal activity.

\medskip
\noindent\textbf{Example 2: Bird Isolation Scene}

\noindent\textbf{Question:} What is the author's intention in presenting a scene where 
all birds fly away except for one on the branch? \\
\textbf{Correct Answer:} C. The bird might get used to being alone.

\medskip

\noindent\textbf{Description 1 (CREST)}

\textit{Part 1 -- Direct Frame Evidence (question-specific)} \\
The scene shows a silhouette of a bird perched on a bare, snow-covered tree branch 
against a gray, overcast sky. The rest of the birds have flown away, leaving only this 
one bird alone. The text overlay reads ``Alone By Edgar Allan Poe,'' indicating that 
the scene references the poem. The contrast between the solitary bird and the empty 
branches emphasizes isolation.

\textit{Part 2 -- Full Domain Coverage (everything related that might be questioned 
later)} \\
The scene conveys themes of solitude and isolation through visual elements such as a 
barren tree, snow-covered branches, and an overcast sky. The reference to Edgar Allan 
Poe suggests a connection to literary themes of loneliness and introspection. The 
imagery of a single bird remaining after others have left reflects persistence, 
emotional isolation, and adaptation to solitude within a natural environment.

\medskip

\noindent\textbf{Description 2 (AKS)}

\textit{Part 1 -- Direct Frame Evidence (question-specific)} \\
The scene shows a barren tree with snow on its branches. Several birds are perched on 
the branches, but one bird is seen flying away while the others remain stationary. The 
text overlay reads ``Alone By Edgar Allan Poe,'' suggesting a theme of solitude.

\textit{Part 2 -- Full Domain Coverage (everything related that might be questioned 
later)} \\
The scene appears to interpret Edgar Allan Poe's poem ``Alone.'' Visual elements such 
as the barren tree, snow-covered branches, and bird activity contribute to a melancholic 
tone. The imagery suggests themes of loneliness, disconnection, and emotional 
reflection. The scene may also relate to broader themes such as mortality, isolation, 
and the human condition.

\medskip

\noindent These examples highlight two key differences: (i) CREST produces descriptions 
that are more tightly aligned with the actual visual evidence and temporal sequence of 
events, and (ii) AKS may introduce inconsistencies or hallucinated elements due to 
missing critical frames, which impacts both factual grounding and downstream reasoning.

\end{document}